\documentclass[]{template}

\usepackage[utf8]{inputenc}             
\usepackage[T1]{fontenc}                
\usepackage{url}                        
\usepackage{booktabs}                   
\usepackage{multirow}
\usepackage{colortbl}
\usepackage{multicol}
\usepackage{amsfonts}                   
\usepackage{nicefrac}                   
\usepackage{microtype}                  
\usepackage[dvipsnames]{xcolor}         
\usepackage[normalem]{ulem}

\usepackage{latexsym}

\usepackage{graphicx}
\usepackage{float}
\usepackage{subcaption}
\usepackage{wrapfig}
\usepackage{lipsum}
\usepackage{geometry}
\usepackage{bm}

\usepackage{tabularx} 
\usepackage{ragged2e} 
\newcolumntype{L}{>{\RaggedRight\hangafter=1\hangindent=0em}X}

\usepackage{enumitem}

\usepackage{amsmath}
\usepackage{amssymb}
\usepackage{mathtools}
\usepackage{amsthm}
\usepackage{siunitx}

\setboolean{logo}{true}    

\usepackage[linesnumbered,ruled,vlined]{algorithm2e}

\hypersetup{
    colorlinks=true,
    linkcolor=red,
    citecolor=Cerulean,
    filecolor=magenta,      
    urlcolor=magenta,
}

\usepackage[capitalize,noabbrev]{cleveref}
\crefname{section}{§}{§§}
\Crefname{section}{§}{§§}

\usepackage{calligra}
\DeclareMathAlphabet{\mathcalligra}{T1}{calligra}{m}{n}

\usepackage{pifont}

\theoremstyle{plain}

\theoremstyle{definition}

\theoremstyle{remark}

\renewcommand{\paragraph}[1]{\vspace{1mm}\noindent\textbf{#1}}

\DeclareCaptionLabelFormat{cont}{#1~#2\alph{ContinuedFloat}}
\newcommand{\eg}{e.g.\@\xspace}

\usepackage[most]{tcolorbox}
\tcbset{
  promptbox/.style={
    top=10pt,
    colback=lightgray!20,
    colframe=Black,
    colbacktitle=NavyBlue,
    enhanced,
    center,
    attach boxed title to top center={yshift=-0.1in,xshift=0.0in},
    boxed title style={boxrule=0pt,colframe=white,},
  }
}
\newtcolorbox{promptbox}[2][]{promptbox, title=#2,#1}
\tcbset{
  takeawaybox/.style={
    top=10pt,
    colback=lightgray!20,
    colframe=Black,
    colbacktitle=BurntOrange,
    enhanced,
    center,
    attach boxed title to top center={yshift=-0.1in,xshift=0.0in},
    boxed title style={boxrule=0pt,colframe=white,},
  }
}
\newtcolorbox{takeawaybox}[2][]{takeawaybox, title=#2,#1}
\tcbset{
  observationbox/.style={
    top=10pt,
    colback=lightgray!20,
    colframe=Black,
    colbacktitle=YellowGreen,
    enhanced,
    center,
    attach boxed title to top center={yshift=-0.1in,xshift=0.0in},
    boxed title style={boxrule=0pt,colframe=white,},
  }
}
\newtcolorbox{observationbox}[2][]{observationbox, title=#2,#1}
\tcbset{
  outlinebox/.style={
    top=10pt,
    colback=yellow!15,
    colframe=orange,
    colbacktitle=orange,
    enhanced,
  }
}
\newtcolorbox{outlinebox}[2][]{outlinebox, title=#2,#1}

\usepackage{xspace}
\newcommand{\modelname}[0]{Kernel-Smith\xspace}

\newcommand\blfootnote[1]{%
  \begingroup
  \renewcommand\thefootnote{}\footnote{#1}%
  \addtocounter{footnote}{-1}%
  \endgroup
}

\usepackage{CJKutf8}

\usepackage{fontawesome5} 
\hypersetup{
    colorlinks=true,
    linkcolor=blue,
    filecolor=magenta,      
    urlcolor=blue,          
}
\title{\modelname: A Unified Recipe for Evolutionary Kernel Optimization}

\author[1,3]{He Du$\dagger$}
\author[1]{Qiming Ge$\dagger$}
\author[1]{Jiakai Hu}
\author[1]{Aijun Yang}
\author[1]{Zheng Cai}
\author[1]{Zixian Huang}
\author[1]{Sheng Yuan}
\author[1]{Qinxiu Cheng}
\author[1]{Xinchen Xie}
\author[1,3]{Yicheng Chen}
\author[1]{Yining Li}
\author[2]{Jiaxing Xie}
\author[2]{Huanan Dong}
\author[2]{Yaguang Wu}
\author[2]{Xiangjun Huang}
\author[2]{Jian Yang}
\author[1]{Hui Wang}
\author[1]{Bowen Zhou}
\author[1]{Bowen Li$\ddagger$}
\author[1]{Qipeng Guo$\ddagger$}
\author[1]{Kai Chen$\ddagger$}
\affil[1]{Shanghai AI Laboratory}
\affil[2]{MetaX}
\affil[3]{Fudan University}

\begin{abstract}
High-performance GPU kernel generation is increasingly important for both large-model systems and broader scientific or industrial workloads, yet current LLM-based approaches still struggle to sustain reliable optimization beyond one-shot code generation. We present \modelname, a framework for high-performance GPU kernel and operator generation that combines a stable evaluation-driven evolutionary agent with an evolution-oriented post-training recipe. On the agent side, \modelname maintains a population of executable candidates and iteratively improves them using an archive of top-performing and diverse programs together with structured execution feedback on compilation, correctness, and speedup. To make this search reliable, we build backend-specific evaluation services for Triton on NVIDIA GPUs and Maca on MetaX GPUs. On the training side, we convert long-horizon evolution trajectories into step-centric supervision and reinforcement learning signals by retaining correctness-preserving, high-gain revisions, so that the model is optimized as a strong local improver inside the evolutionary loop rather than as a one-shot generator. Under a unified evolutionary protocol, \modelname-235B-RL achieves state-of-the-art overall performance on KernelBench with Nvidia Triton backend, attaining the best average speedup ratio and outperforming frontier proprietary models including Gemini-3.0-pro and Claude-4.6-opus. We further validate the framework on the MetaX MACA backend, where our \modelname-MACA-30B surpasses large-scale counterparts such as DeepSeek-V3.2-think and Qwen3-235B-2507-think, highlighting potential for seamless adaptation across heterogeneous platforms. Beyond benchmark results, the same workflow produces upstream contributions to production systems including SGLang and LMDeploy, demonstrating that LLM-driven kernel optimization can transfer from controlled evaluation to practical deployment.
\end{abstract}

\begin{document}

\maketitle

\vspace{-1em}
\begin{center}
    \begin{tabular}{l l}
        \faGlobe \ \textbf{Project Page:} & \href{https://chat.intern-ai.org.cn/kernel-smith}{\texttt{https://chat.intern-ai.org.cn/kernel-smith}$^*$} \\
        \faDatabase \ \textbf{Data Access:} & Please contact us via E-mail for access.
    \end{tabular}
\end{center}
\vspace{-3em}

\blfootnote{$^*$ The online demo is powered by a customized version of the Intern-S1-Pro model \cite{zou2026interns1proscientificmultimodalfoundation}.}
\blfootnote{$\dagger$ These authors contributed equally to this work.}
\blfootnote{$\ddagger$ Corresponding authors: Bowen Li (libowen@pjlab.org.cn), Qipeng Guo (guoqipeng@pjlab.org.cn) Kai Chen (chenkai@pjlab.org.cn)}

\section{Introduction}

High-performance kernels are central to translating hardware capability into practical throughput on modern accelerators. Systems such as Megatron~\cite{shoeybi2019megatron}, XTuner~\cite{xtuner2023}, vLLM~\cite{kwon2023efficient}, SGLang~\cite{zheng2023sglang}, and LMDeploy~\cite{lmdeploy2023} have demonstrated that careful kernel optimization can improve large-model training and inference by large margins. This dependence on kernel engineering extends well beyond foundation models: scientific computing workloads in AI for Science (AI4S)~\cite{zhang2023scientific} and deployment pipelines in diverse industrial settings likewise rely on efficient operator implementations to realize the performance potential of the underlying hardware.

Although programming has become a representative capability of modern LLMs~\cite{roziere2023code, chen2021evaluating}, recent studies suggest that high-performance kernel generation remains far from solved~\cite{ouyang2025kernelbench, wen2025multikernelbench, zhu2026cudabench, guantritongym}. In particular, achieving end-to-end autonomous contributions to real production repositories is still highly challenging. We argue that making LLM-based kernel development practical requires solving two coupled problems. First, efficient kernels usually emerge only after searching over many implementation choices, including alternative fusion patterns, tiling strategies, and rewrite directions. Existing systems increasingly rely on multi-turn refinement or history-conditioned agent loops~\cite{wei2025astra, zhang2025cudaforge, lei2025pragma, baronio2507kevin, liu2026drkernel, dai2026cuda}. While useful for localized debugging, these procedures can anchor later proposals to early decisions and limit exploration diversity. Second, functional correctness and high performance are not the same capability. The objective is therefore not merely to generate one correct and fast kernel in a single pass, but to sustain iterative optimization that keeps improving candidate programs and makes effective use of additional test-time compute.

To address these challenges, we propose \modelname, a unified framework that combines a reliable evaluation-driven evolutionary agent with a training recipe tailored to evolutionary search through the identification of key improvement steps from evolution trajectories.
An overview of the full framework is shown in Figure~\ref{fig:main_fig}.

The first design choice of \modelname lies in its agent framework. Evolutionary search is a natural fit for kernel optimization because it maintains a population of executable candidates and allows performance gains to accumulate over multiple rounds of search~\cite{novikov2025alphaevolve}. However, this paradigm is highly sensitive to evaluation variance: when profiling noise is large, the search may preserve suboptimal kernels or eliminate genuinely promising ones, and such mistakes compound across generations. We therefore center the agent design on kernel-specific evaluation stability, combining fixed computation graphs, repeated measurements, and outlier removal to suppress timing noise and preserve reliable search dynamics.

The second design choice of \modelname lies in its training recipe. Rather than optimizing the model for one-shot kernel generation, we train it to act as a strong local improver inside the evolutionary loop. Concretely, we transform long-horizon evolution trajectories into step-centric training signals and retain only the high-gain revisions that move a candidate toward better correctness-preserving performance. This filtering strategy acts as a form of trajectory compression: instead of imitating every intermediate transition, the model learns the atomic improvements that contribute most to eventual speedup. We apply the same principle in both supervised fine-tuning and reinforcement learning, where carefully selected optimization steps provide more informative learning signals than full trajectories that may contain redundant transitions or shortcut opportunities. As a result, \modelname improves not only single-step edit quality, but also the rate at which gains compound over successive rounds of evolutionary search.

\begin{figure}[htbp]
    \centering
    \includegraphics[width=\textwidth]{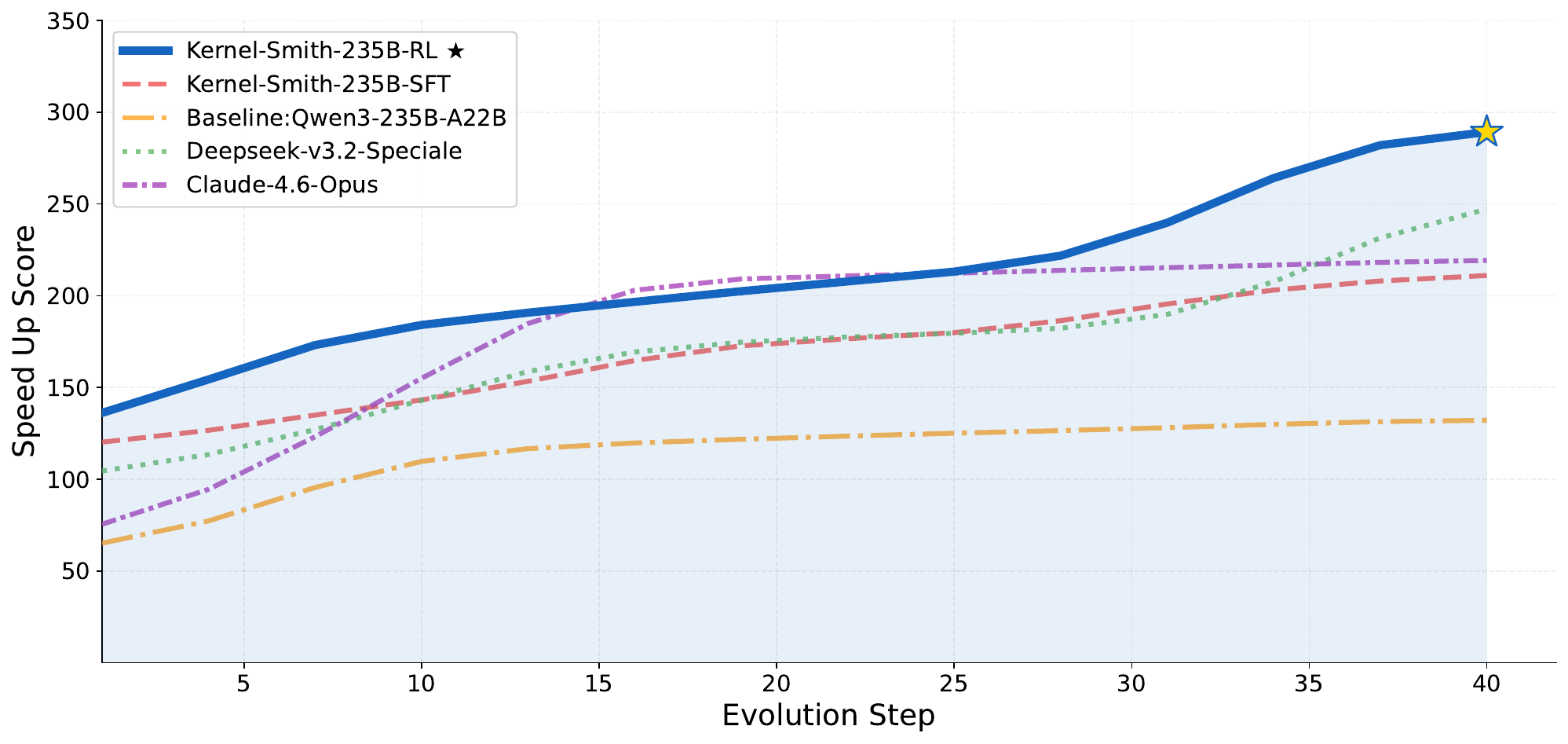}
    \caption{Comparison of the best program score trajectories across various models over successive evolutionary generations. The program score is formulated as a linear function directly proportional to the speedup, penalizing programs that fail compilation or correctness checks by assigning them appropriately low scores. Throughout the evolution steps, our RL model (\modelname-235B-RL) exhibited superior efficiency for evolution and peak performance.}
    \label{fig:best_score}
\end{figure}

These two design choices translate into clear empirical gains. Under a unified evolutionary-agent protocol, \modelname-235B-RL achieves state-of-the-art overall performance on KernelBench~\cite{ouyang2025kernelbench}, attaining the best average speedup ratio while maintaining strong correctness against competitive open-weights baselines as well as frontier proprietary models such as Gemini-3.0-pro and Claude-4.6-opus. More importantly, Figure~\ref{fig:best_score} shows that its best-score growth curve forms the upper envelope of competing models throughout the search process, indicating that our model benefits more effectively from additional test-time compute. This result directly reflects the role of our two core components: stable evaluation preserves reliable search dynamics, while evolution-oriented post-training improves the quality of each optimization step and allows gains to compound over longer horizons. Beyond benchmark performance, we further validate the practical value of \modelname through accepted pull requests to widely used inference engines including SGLang \cite{zheng2023sglang} and LMDeploy \cite{lmdeploy2023}, demonstrating that the framework transfers from controlled evaluation to real deployment settings.

\section{Related Work}

\subsection{Benchmarks for LLM-Driven Kernel Generation}
Benchmark design has become central to LLM-driven kernel generation because pass rate alone does not capture whether a generated kernel is actually useful in practice. KernelBench \cite{ouyang2025kernelbench} established the canonical evaluation setting by formulating the task as replacing PyTorch reference implementations with faster GPU kernels and by introducing the $\mathrm{fast_p}$ family of metrics, which jointly reflects correctness and speedup. Subsequent benchmarks extend this setup along complementary axes rather than simply increasing scale. MultiKernelBench \cite{wen2025multikernelbench} broadens evaluation beyond a single hardware stack to study cross-platform kernel generation, CUDABench \cite{zhu2026cudabench} expands the task scope toward text-to-CUDA generation , and TritonGym \cite{guantritongym} focuses on benchmarking agentic workflows for Triton code generation. Taken together, these benchmarks move the field from anecdotal case studies toward reproducible, execution-grounded evaluation, while also underscoring that strong results on standardized tasks do not yet fully resolve the challenges of heterogeneous and production-facing kernel optimization.


\subsection{Agent Systems and Model Training for Kernel Generation}
A critical bottleneck in automating kernel generation is the scarcity of human-optimized, high-performance CUDA and Triton code, which limits standard supervised fine-tuning. To break this ceiling, recent advancements utilize Reinforcement Learning from Verifiable Rewards (RLVR) \cite{tehrani2026fine}. For instance, AutoTriton \cite{li2025autotriton} combines an automated data distillation pipeline with Group Relative Policy Optimization (GRPO), utilizing rule-based and execution-based rewards to specifically establish foundational Triton programming capabilities.
Beyond single-pass generation, multi-agent workflows and multi-turn RL are designed to replicate the iterative debugging and tuning trajectories of performance engineers. Frameworks such as Astra and CudaForge partition the cognitive load into specialized roles, iteratively refining kernels based on feedback from profilers like NVIDIA Nsight Compute (NCU) \cite{wei2025astra, zhang2025cudaforge}. PRAGMA \cite{lei2025pragma} further advances this by injecting fine-grained hardware metrics into a bottleneck-aware reasoning module. However, training models to natively perform this iterative optimization introduces unique RL challenges such as context explosion and sparse reward attribution. Kevin \cite{baronio2507kevin} addresses these by formulating a multi-turn RL recipe that effectively evaluates and attributes rewards to intermediate refinement turns. Dr. Kernel \cite{liu2026drkernel} further identifies gradient biases in multi-turn advantage estimation and introduces Turn-level Reinforce-Leave-One-Out (TRLOO), alongside Profiling-based Rewards (PR) and Rejection Sampling (PRS) to mitigate prevalent issues like reward hacking and "lazy optimization" (e.g., only fusing trivial operations). Scaling these concepts, CUDA Agent \cite{dai2026cuda} proposes a comprehensive agentic RL system featuring combinatorial data synthesis and stable multi-stage warm-up, achieving substantial speedups over industrial compilers like TorchInductor across varied difficulty levels.

\subsection{Advanced Search and Evolution Algorithms}
Because the GPU kernel optimization landscape is highly non-convex, recent work increasingly treats kernel generation as a structured search problem rather than a pure one-shot prediction task. KernelSkill addresses repetitive backtracking with a dual-level memory architecture that retrieves previously verified optimization skills \cite{sun2026kernelskillmultiagentframeworkgpu}. KernelBand instead emphasizes exploration--exploitation balance, formulating optimization as a hierarchical multi-armed bandit that uses runtime behavior to prune unpromising branches \cite{ran2025kernelband}. K-Search pushes this perspective further by co-evolving high-level algorithmic planning and low-level implementation, replacing blind code mutation with search over a more explicit world model of hardware-software interaction \cite{cao2026k}.
Complementary lines of work move this search process into the learning objective itself. CUDA-L1 \cite{li2025cuda} introduces contrastive reinforcement learning, conditioning policy updates on multiple previously generated code variants and their measured speedups so that the model can reason more explicitly about performance trade-offs. CUDA-L2 \cite{su2025cuda} scales this idea to large HGEMM optimization spaces and shows that reinforcement learning can be used as a targeted search procedure over highly specialized kernel families. At an even more aggressive end of the spectrum, TTT-Discover \cite{yuksekgonul2026learning} performs reinforcement learning only at test time for a single problem, extending the search horizon for difficult scientific discovery tasks. Taken together, these approaches suggest that progress in kernel generation depends not only on stronger base models or richer feedback, but also on search algorithms that better organize exploration across candidate implementations.
\section{\modelname}


\subsection{Overview}
Our task is to generate high-performance GPU kernels from PyTorch reference operators. Given a PyTorch module together with its execution interface and test inputs, \modelname produces candidate kernel implementations whose goal is not only to preserve functional behavior, but also to improve execution efficiency on the target hardware. The target therefore combines three requirements: each candidate must compile successfully, match the numerical output of the PyTorch reference, and deliver measurable speedup over the eager-mode baseline.

To address this objective, \modelname adopts an evolve-agent framework rather than the conventional multi-turn agent loop used in prior kernel optimization systems \cite{wei2025astra,zhang2025cudaforge,lei2025pragma}. Instead of refining a single trajectory through sequential dialogue, the system maintains and evolves a population of candidate programs, which broadens exploration over the kernel search space and better exploits test-time compute. This evolution process is paired with a comprehensive automated evaluation backend that executes generated kernels, returns structured feedback, and measures compilation, correctness, and speedup in a stable and reliable manner. 

\begin{figure}[htbp]
    \centering
    \includegraphics[width=\textwidth]{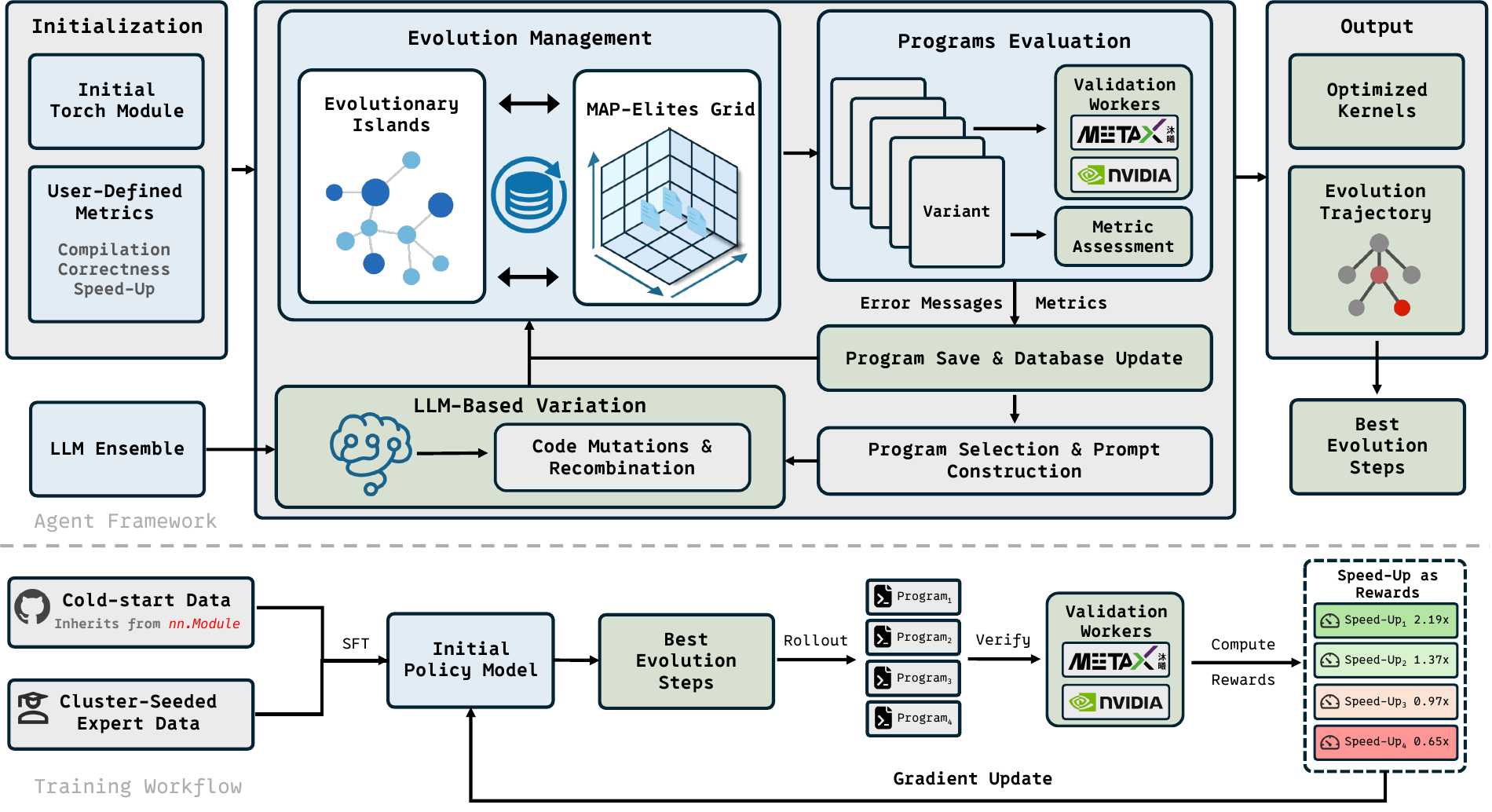}
    \caption{An overview of the evolutionary process of our proposed framework and the corresponding model training pipeline.}
    \label{fig:main_fig}
\end{figure}

\subsection{Agent Framework}
AlphaEvolve formulates code optimization as an evolutionary search process over executable programs: the model proposes candidate edits, an evaluator scores the resulting programs, and the search state is maintained in an archive that supports subsequent exploration \cite{novikov2025alphaevolve}. This perspective is especially well suited to machine-verifiable tasks such as kernel generation, where correctness and performance can both be measured automatically. More broadly, the design is closely related to island-based evolutionary algorithms, which preserve partially independent search trajectories, and MAP-Elites, which maintains diverse high-quality solutions across a feature space rather than collapsing the search to a single incumbent \cite{whitley1999island,mouret2015illuminating}.

We instantiate this idea through OpenEvolve\footnote{\url{https://github.com/algorithmicsuperintelligence/openevolve}}, adapting an evolutionary coding agent to the setting of high-performance GPU kernel generation. In our setting, each search state corresponds to a backend-specific kernel candidate for a fixed PyTorch reference module. At each iteration, the agent is prompted with the reference implementation together with archived candidates sampled from both top-performing and diverse regions of the search space, and then proposes a new kernel implementation. Following our design, the archive is organized by a feature space that includes kernel complexity and an overall score combining compilation, correctness, and speedup. 
Our main adaptation to kernel generation is a fine-grained execution feedback mechanism at every evolution step. Rather than returning only a scalar reward, the evaluator produces structured feedback that includes compilation status, correctness outcomes, speedup, runtime measurements, hardware metadata, and error logs. These signals are injected into the next iteration together with archived candidate programs, allowing the model to learn not only from strong solutions but also from informative failure cases. As a result, the agent performs iterative kernel optimization with explicit execution evidence instead of relying purely on conversational refinement. A representative example of the system prompt, user prompt, and model generation for one evolution step is provided in Appendix~\ref{app:evolve-prompt-example}.

\subsection{Evaluation Backends}

We established a comprehensive automated evaluation system designed to multi-dimensionally verify the reliability and acceleration effects of generated GPU operators in high-performance computing scenarios.

\paragraph{Evaluation Service and Metrics}
A distributed API evaluation server was developed to provide distributed parallel evaluation interfaces. In our current backends, the system generates Triton kernels for NVIDIA GPUs and Maca kernels for MetaX GPUs. The core evaluation metrics include: 1) Compilation, which verifies whether the generated backend-specific code can be successfully compiled on the target hardware; 2) Correctness, which examines the numerical consistency between the operator output and the PyTorch reference implementation; and 3) Speedup, which measures the performance improvement of the generated operators relative to the PyTorch eager mode.

\paragraph{Stability and Noise Reduction}
In GPU environments, the wall-clock time of operator execution exhibits non-negligible fluctuations even when hardware and driver versions are fixed. For small-scale input tensors, the kernel launch time accounts for an excessive proportion of total execution, leading to particularly pronounced volatility. 
To mitigate these effects, we implemented the following solutions: first, warm-up executions were performed before timing to reduce initialization overhead and transient variance; second, multiple measurements were conducted to calculate the mean and exclude outliers; third, CUDAGraph technology was introduced to further stabilize the timing process. The improved evaluation scheme successfully constrained execution time fluctuations to within $1\%$.

\paragraph{Hacking Detection}
In automated generation tasks, models may circumvent backend-specific kernel generation by directly calling native PyTorch operators, thereby fabricating a false "passed test" result with approximately 1x speedup. 
We established a runtime detection mechanism to mandate the actual execution of generated kernel code rather than falling back to PyTorch implementations. 
Beyond such automatically detectable cases, we also manually observed a failure mode in strong closed-source models that we refer to as \emph{advanced hacking} or \emph{trivial optimization}. In these cases, the model applies optimizations that satisfy compilation and correctness checks but offer little practical engineering value; rewriting simple element-wise additions in Triton or Maca is one representative example. This behavior is closely related to the lazy optimization phenomenon discussed in Dr.~kernel \cite{liu2026drkernel}.

\paragraph{Heterogeneous Platforms}
As illustrated in Figure~\ref{fig:main_fig}, our evaluation backend follows a backend-decoupled design that separates task specification, execution orchestration, and metric computation from device-specific compilation and runtime interfaces, rather than being tied to a single vendor-specific stack.  This allows the same evaluation protocol to be reused across heterogeneous accelerators while preserving a consistent optimization target for the agent. In the current implementation, we instantiate this design with Triton backends for NVIDIA GPUs and MACA backends for MetaX GPUs, both evaluated under the same compilation, correctness, and speedup criteria. The same abstraction also provides a natural extension path to additional platforms, such as Huawei NPUs, without changing the agent-side optimization objective.


\section{Training Recipe}

\subsection{Overview}

Our training recipe is designed to improve the model's effectiveness inside the evolutionary agent rather than optimize one-shot kernel generation in isolation. Starting from a curated corpus of PyTorch modules, we synthesize multi-step evolution trajectories with strong teacher models and convert them into post-training signals at the level of individual improvement steps. This step-centric formulation treats multi-round search as a composition of learnable atomic revisions: supervised fine-tuning provides a cold start from correctness- and speedup-filtered samples, while reinforcement learning further sharpens the model on the most informative high-gain steps.

\subsection{Data Synthesis}

\paragraph{Torch Data Curation}
Existing work typically starts from fixed benchmarks or a small set of well-maintained libraries, and may further increase complexity by synthetically combining simple operators into fused tasks. These pipelines are effective for constructing training problems at scale, but their seed distributions still remain biased toward canonical operators and standardized repository structures, leaving limited coverage of the diverse implementation patterns found in real-world codebases. 
To address this gap, we systematically crawl diverse GitHub repositories and build an automated static-analysis pipeline to extract \texttt{torch.nn.Module} subclasses from wild code. 

More specifically, we first filter high-quality open-source GitHub repositories to obtain a diverse pool of PyTorch implementations. We then extract candidate \texttt{nn.Module} definitions and, rather than discarding incomplete files, recursively resolve intra-file dependencies, inline essential components, and infer the minimal PyTorch imports needed to make each example self-contained. After this normalization step, we apply embedding- and graph-based deduplication to reduce near-duplicate modules while preserving structural diversity. We further use LLM-assisted test generation to supplement missing test cases, followed by execution-based filtering to remove examples that fail to run reliably. Ultimately, this pipeline yields 59k high-quality modules spanning 20 functional families, forming a robust PyTorch dataset.

\paragraph{Instruction Data Synthesis}
We leverage our \modelname framework for data synthesis to specifically address the challenges of data diversity and quality. Our synthesis process is divided into two primary components:

\begin{itemize}
    \item \textbf{Cold-start Data}: Starting from the curated PyTorch dataset, we run \modelname with the open-source teacher model DeepSeek-V3.2-Speciale \citep{liu2025deepseek} to generate rollout trajectory data. We then filter these trajectories using correctness and speedup metrics, retaining only samples that are both functionally valid and performance-improving.
    
    \item \textbf{Cluster-Seeded Expert Data}: To raise the quality ceiling of the synthesized data, we first embed the curated dataset and then apply HDBSCAN clustering \citep{mcinnes2017hdbscan} to identify representative cluster centers for manual cleaning and expert annotation. We then feed these expert-curated operators back into \modelname for additional rollout rounds, producing higher-fidelity trajectory data with stronger overall performance.
\end{itemize}


\subsection{Supervised Fine-tuning}

We utilize seed-based synthetic data for the SFT cold-start phase. Our pipeline is structured around multi-turn agentic evolution tasks, where we preserve the historical state of each step and decompose multi-turn trajectories into single-turn training samples. Given that operator generation requires both functional correctness and performance gains, we implement the following dual-filtering strategies:
\begin{itemize}
    \item \textbf{Correctness-Oriented Augmentation} This stage focuses on the model's fundamental generation accuracy. We specifically filter the initial translation step (PyTorch $\to$ Triton) and adopt a relaxed filtering policy to include all functionally correct outputs. This ensures the model's proficiency in basic code translation.
    \item \textbf{Performance-Oriented Augmentation} For the iterative evolution phase (Triton $\to$ Triton), we enforce a more stringent filtering policy. Only samples that are both functionally correct and achieve a speedup ratio > 1.0 are selected, thereby enhancing the model's iterative optimization capabilities.
\end{itemize}
To ensure a balanced training distribution, we categorize operator difficulty using heuristic rules based on the number and types of modules involved. Finally, we perform balanced sampling across these categories, resulting in over 200k high-quality single-turn samples for SFT training with a 64k context length.

\subsection{Reinforcement Learning}

The highly complex and nonlinear nature of the entire evolution process poses significant challenges for end-to-end on-policy reinforcement learning training. Consequently, the core problem shifts toward identifying the most effective single-step evolution RL strategy to maximize generalization performance across the multi-round evolution process during inference. We investigated several strategies for selecting evolution steps and derived the following empirical observations:

\paragraph{The incorporation of all procedural steps may facilitate the emergence of shortcuts derived from information leakage.} Incorporating all steps from the evolution process significantly expands the training set. However, experimental results indicate that including both preceding and succeeding steps simultaneously allows the model to exploit the presence of superior kernel examples within the input prompts of later steps. Consequently, the model tends to memorize these high-quality references rather than acquiring generalized optimization capabilities. Although this leads to an ostensibly favorable reward curve, the actual learning efficacy remains marginal.

\paragraph{Selecting only the initial step of the evolution process yields suboptimal performance.} Our analysis suggests that the distribution of the first step deviates substantially from subsequent stages, as the input consists solely of a PyTorch implementation without any optimized Triton kernels as exemplars. Furthermore, the primary objective of the initial step is the functional migration from PyTorch to Triton rather than achieving substantial throughput acceleration. The inherent simplicity of this task renders it unsuitable for effective reinforcement learning.

\paragraph{Selecting the best steps from the evolution process results in a marked improvement in performance.}
The inclusion of best steps typically involves providing example code with a certain baseline level of acceleration. Leveraging such inputs to generate further optimized kernels ensures that the learning space for the model remains constrained while maintaining a sufficient level of challenge. During a single training iteration, the steady increase in the reward curve indicates that the task complexity is appropriately calibrated. Furthermore, consistent end-to-end performance improvements observed across multiple rounds of inference demonstrate that the best step effectively represents the fundamental atomic capability within the iterative evolutionary process.

As illustrated in Figure~\ref{fig:main_fig}, our methodology is based on Cluster-Seeded Expert Data, where each problem undergoes 40 iterations of evolutionary refinement using Gemini-3.0-pro. From this process, the Best steps are selected to construct the final training set. A representative data sample comprises an input containing several high-performance kernels as exemplars and a designated parent code for modification. The model is then tasked with generating a more efficient and optimized kernel based on these inputs; the detailed configurations are provided in the prompt template in Appendix~\ref{app:evolve-prompt-example}. We employ the GRPO algorithm, sampling eight candidates per data entry and utilizing the speedup ratio relative to the parent code as the reward signal for training. The results presented in Table~\ref{tab:horizontal_benchmark} demonstrate the efficacy of our proposed training strategy.

\section{Experiments}

\subsection{Evaluation Protocol}
\paragraph{Models}
To comprehensively evaluate the effectiveness of our proposed approach, we conduct a rigorous comparison against a diverse set of state-of-the-art baselines. For the open-weights baselines, we select the recent reasoning-focused Qwen3 series~\cite{yang2025qwen3} (including both Qwen3-235B-A22B-2507-think and Qwen3.5-397B-A17B-think) to assess performance across varying parameter scales. We also include the highly competitive DeepSeek-v3.2-Speciale~\cite{liu2025deepseek}, alongside recently open-sourced frontier models such as Kimi-K2.5~\cite{team2026kimi} and MiniMax-M2.5~\cite{minimax2026m25}.
Furthermore, to establish the absolute performance upper bound and ensure a holistic evaluation, we additionally benchmark against leading proprietary systems, specifically Claude-4.6-opus and Gemini-3.0-pro.

\paragraph{Configurations}
To ensure a rigorous and fair evaluation, we \textit{deploy all compared models within the same \modelname evolution-agent framework}, making the agent system itself a controlled constant across the entire comparison. Under this unified protocol, we conduct 40 rounds of iterative evolution for each model. For text generation, the decoding parameters are set to a temperature of 0.6 and a top-$p$ of 0.95. To prevent context overflow during extended interactions, we strictly cap both the input prompt and the maximum output generation at 32K tokens per round. Furthermore, to ensure a rigorous assessment of system stability and mitigate variance, we execute independent unit tests 100 times for each individual module and report the average performance.

\paragraph{Metrics}
To rigorously assess the performance of the generated operators, we evaluate the models based on three primary metrics:
\begin{itemize}
    \item \textbf{Correctness (corr)}: Measures the validity of the generated operators after the introduction of hack detection. An operator is considered correct only if its computational precision difference compared to the original implementation is strictly controlled within an acceptable threshold.
    \item \textbf{Fast Proportion ($\mathbf{fast_p}$)}: Represents the percentage of generated operators that successfully achieve execution speedup (i.e., faster execution time) compared to the original baseline operators.
    \item \textbf{Average Speedup Ratio (avg amsr)}: Calculates the mean speedup ratio across all operators within a specific difficulty level. Specifically, to emphasize acceleration performance, all instances where the speedup ratio was less than 1 were assigned a score of zero. This serves as the core indicator of the absolute performance gains delivered by the generated code.
\end{itemize}

\subsection{Results On Nvidia Backend}

Table \ref{tab:horizontal_benchmark} presents the performance of our proposed model alongside leading open-weights models and powerful proprietary baselines. As expected, the highly optimized proprietary model, Claude-4.6-opus, establishes a performance upper bound in overall correctness (corr of 99.33) and first-pass speed ($\mathrm{fast_1}$ of 0.77). However, our approach demonstrates highly competitive accuracy, achieving an average corr of 96.33 to effectively outperform all other advanced baselines, including Gemini-3.0-pro (94.33) and DeepSeek-v3.2-Speciale (94.67). More importantly, our model establishes a new state-of-the-art in overall generation quality and structural stability, recording the highest avg amsr of 3.70 across all difficulty levels. This distinct advantage is particularly pronounced in the Level 2 subset, where our model delivers a remarkable avg amsr of 7.77, substantially exceeding even Claude-4.6-opus (5.83). Furthermore, on the hardest tasks (Level 3), our model sustains a robust correctness rate of 94, surpassing Gemini-3.0-pro and all open-weights counterparts by significant margins. These findings suggest that while ultra-large proprietary models may lead in raw accuracy, our method successfully optimizes for deeper reasoning (avg amsr), offering an exceptionally robust alternative for complex tasks.

\begin{table*}[htbp]
\centering
\caption{Benchmark Results with Baselines and Our Model. \textbf{Best} results are in bold, and \uline{second best} are underlined.}
\label{tab:horizontal_benchmark}
\resizebox{\textwidth}{!}{%
\begin{tabular}{l ccc ccc ccc ccc}
\toprule
\multirow{2}{*}{\textbf{Models}} & \multicolumn{3}{c}{\textbf{Level 1 (Easy)}} & \multicolumn{3}{c}{\textbf{Level 2 (Medium)}} & \multicolumn{3}{c}{\textbf{Level 3 (Hard)}} & \multicolumn{3}{c}{\textbf{AVG Level-1/2/3}} \\
\cmidrule(lr){2-4} \cmidrule(lr){5-7} \cmidrule(lr){8-10} \cmidrule(lr){11-13}
& Corr & $\mathrm{Fast_1}$ & Avg AMSR & Corr & $\mathrm{Fast_1}$ & Avg AMSR & Corr & $\mathrm{Fast_1}$ & Avg AMSR & Corr & $\mathrm{Fast_1}$ & Avg AMSR \\

\midrule
\multicolumn{13}{l}{\textit{Baselines}} \\
\hspace{1em} Qwen3-235B-2507-think  & 92 & 0.57 & 1.86 & 96 & 0.86 & 3.96 & 84 & 0.42 & 0.76 & 90.67 & 0.62 & 2.20 \\
\hspace{1em} Qwen3.5-397B-think     & 89 & 0.55 & 2.00 & 90 & 0.72 & 3.86 & 74 & 0.38 & 0.61 & 84.33 & 0.55 & 2.16 \\
\hspace{1em} Minimax-M2.5           & 70 & 0.49 & \uline{2.39} & 91 & 0.77 & 1.06 & 56 & 0.30 & 0.36 & 72.33 & 0.52 & 1.27 \\
\hspace{1em} Kimi-K2.5              & 89 & 0.55 & 2.00 & 90 & 0.72 & 3.86 & 74 & 0.38 & 0.61 & 84.33 & 0.55 & 2.16 \\
\hspace{1em} DeepSeek-v3.2-Speciale & 98 & 0.63 & 2.30 & 96 & 0.88 & \uline{6.89} & 90 & 0.32 & 1.14 & 94.67 & 0.61 & \uline{3.44} \\
\hspace{1em} Gemini-3.0-pro         & \uline{99} & \textbf{0.78} & \textbf{2.46} & 96 & \uline{0.95} & 4.78 & 88 & \uline{0.50} & \uline{1.26} & 94.33 & \uline{0.74} & 2.83 \\
\hspace{1em} Claude-4.6-opus        & \textbf{100}& \uline{0.70} & 2.14 & \textbf{100}& \textbf{0.99} & 5.83 & \textbf{98} & \textbf{0.62} & \textbf{2.02} & \textbf{99.33}& \textbf{0.77}& 3.33\\

\midrule
\multicolumn{13}{l}{\textit{Ours}} \\
 \hspace{1em} \modelname-235B-RL & 97 & \uline{0.70} & 2.30 & \uline{98} & 0.93 & \textbf{7.77} & \uline{94} & 0.46 & 1.02 & \uline{96.33} & 0.70 & \textbf{3.70}\\

\bottomrule
\end{tabular}%
}
\end{table*}

\subsection{Results On MetaX Backend}

We further validate our system on the MetaX platform. In this setting, the task formulation differs slightly from that on the NVIDIA platform: given a CUDA operator implementation, the system is tasked to generate a corresponding high-performance MACA implementation. We construct an evaluation benchmark consisting of 45 common operators, categorized into four groups: Activation (15,~\eg, ReLU, Softmax), Normalization (8,~\eg, BatchNorm, GroupNorm), Reduction\&Aggregation (17,~\eg, Pooling, Sum), and Loss Function (5,~\eg, CrossEntropy, MSE). Each test sample includes a correctness-verified CUDA implementation generated by an LLM, which serves as the input for evolution and as the reference for calculating the speedup ratio. Using a similar agentic framework and evaluation protocol, we compare \modelname-MACA-30B with multiple baselines. Results are shown in Table~\ref{tab:metax_results}. Since the task input includes a correctness-verified CUDA reference, all evaluated models exhibit stable performance on the corr and $\mathrm{fast_1}$ metrics. On the avg asmr metric, our \modelname-MACA-30B outperforms large-scale baselines such as DeepSeek-v3.2-think and Qwen3-235B-2507-think, and \modelname-MACA-235B achieves additional performance gains, highlighting the potential of the proposed framework for generating high-performance operators across heterogeneous platforms.

\begin{table*}[htbp]
\centering
\caption{Evaluation results on MACA kernel generation task.}
\label{tab:metax_results}
\resizebox{\textwidth}{!}{%
\begin{tabular}{l ccc ccc ccc ccc ccc}
\toprule
\multirow{2}{*}{\textbf{Models}} & \multicolumn{3}{c}{\textbf{Activation}} & \multicolumn{3}{c}{\textbf{Normalization}} & \multicolumn{3}{c}{\textbf{Reduction\&Aggregation}} & \multicolumn{3}{c}{\textbf{Loss Function}} & \multicolumn{3}{c}{\textbf{AVG}}\\
\cmidrule(lr){2-4} \cmidrule(lr){5-7} \cmidrule(lr){8-10} \cmidrule(lr){11-13} \cmidrule(lr){14-16}
& Corr & $\mathrm{Fast_1}$ & Avg AMSR & Corr & $\mathrm{Fast_1}$ & Avg AMSR & Corr & $\mathrm{Fast_1}$ & Avg AMSR & Corr & $\mathrm{Fast_1}$ & Avg AMSR & Corr & $\mathrm{Fast_1}$ & Avg AMSR \\
\midrule
\multicolumn{13}{l}{\textit{Baselines}} \\
\quad GPT-OSS-20B          & 100 & 1.00 & 3.25 & 100 & 0.88 & 29.7 & 94.1 & 0.53 & 3.63 & 100 & 0.80 & \textbf{8.56} & 97.8 & 0.78 & 8.69 \\
\quad Qwen3-30B-A3B          & 100 & 1.00 & 9.03 & 100 & 1.00 & 7.03 & 100 & 0.71 & \textbf{17.44} & 100 & 0.80 & 5.67 & 100 & 0.87 & 11.48 \\
\quad Qwen3-235B-2507-think          & 100 & 1.00 & \textbf{14.55} & 100 & 0.75 & 35.18 & 100 & 0.59 & 1.36 & 100 & 1.00 & \uline{6.12} & 100 & 0.80 & 12.30 \\
\quad DeepSeek-v3.2-think          & 100 & 1.00 & 10.09 & 100 & 0.88 & 6.06 & 94.1 & 0.47 & 8.60 & 100 & 0.60 & 2.89 & 97.8 & 0.73 & 8.01 \\
\quad Kimi-K2.5          & 100 & 1.00 & 12.62 & 100 & 1.00 & 29.37 & 100 & 0.65 & 5.11 & 100 & 0.60 & 2.21 & 100 & 0.82 & 11.60 \\
\midrule
\multicolumn{13}{l}{\textit{Ours}} \\
\hspace{1em} \modelname-MACA-30B                & 100 & 1.00 & \uline{13.61} & 100 & 0.88 & \uline{36.03} & 100 & 0.71 & 4.69 & 100 & 0.40 & 5.02 & 100 & 0.80 & \uline{13.27} \\
\hspace{1em} \modelname-MACA-235B                & 100 & 1.00 & 9.25 & 100 & 0.88 & \textbf{40.59} & 100 & 0.65 & \uline{9.63} & 100 & 1.00 & 3.07 & 100 & 0.84 & \textbf{14.26} \\
\bottomrule
\end{tabular}%
}
\end{table*}

\begin{subappendices}

\end{subappendices}

\section{Real-world Applications}

Beyond benchmark evaluation, we are interested in whether the same optimization workflow can transfer to multiple realistic deployment settings. We therefore highlight three complementary case types. The LMDeploy and SGLang cases show that \modelname can contribute to production inference engines through merged upstream pull requests. The Engram case instead focuses on an up-to-date PyTorch module collected from the recent research from DeepSeek \cite{cheng2026conditional}, which helps reduce the risk of benchmark contamination while testing the method on a practically relevant target.

\begin{figure}[htbp]
    \centering
    \begin{subfigure}[b]{0.32\textwidth}
        \centering
        \includegraphics[width=\textwidth]{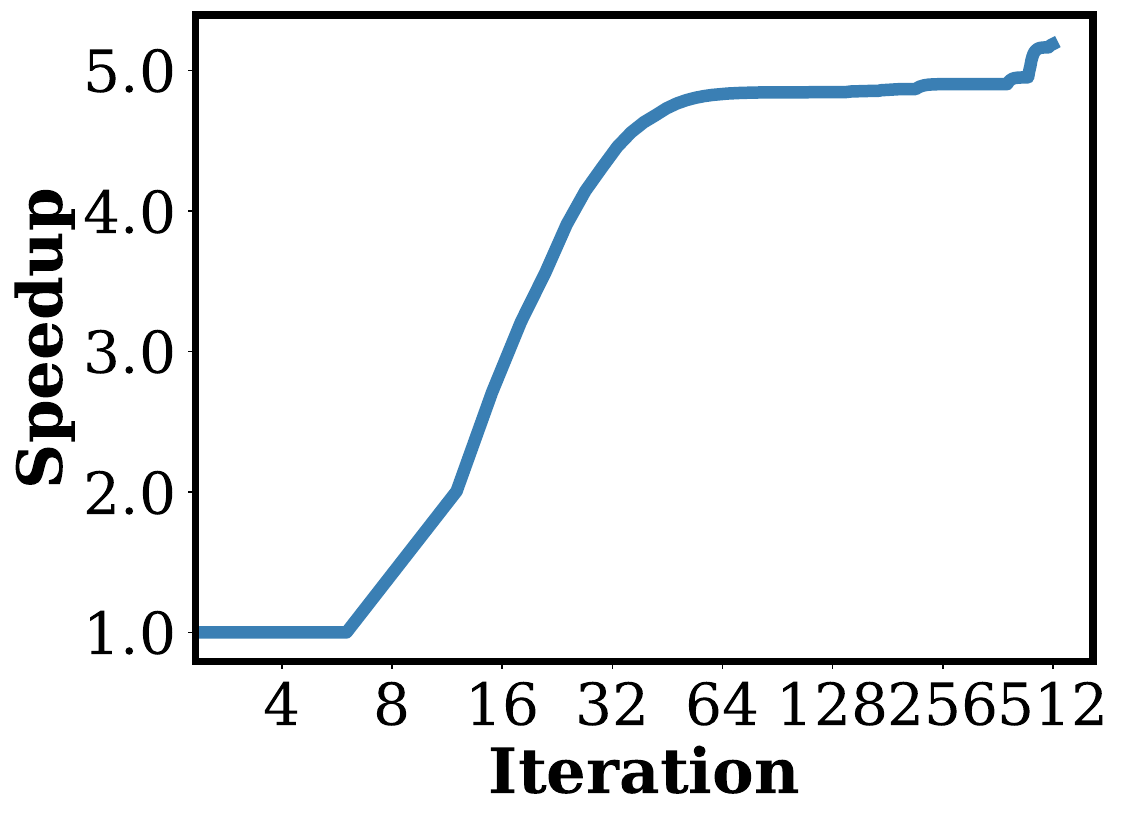}
        \captionsetup{justification=centering}
        \caption{SGLang Normal Decode Set Meta}
        \label{fig:sub1}
    \end{subfigure}
    \hfill
    \begin{subfigure}[b]{0.32\textwidth}
        \centering
        \includegraphics[width=\textwidth]{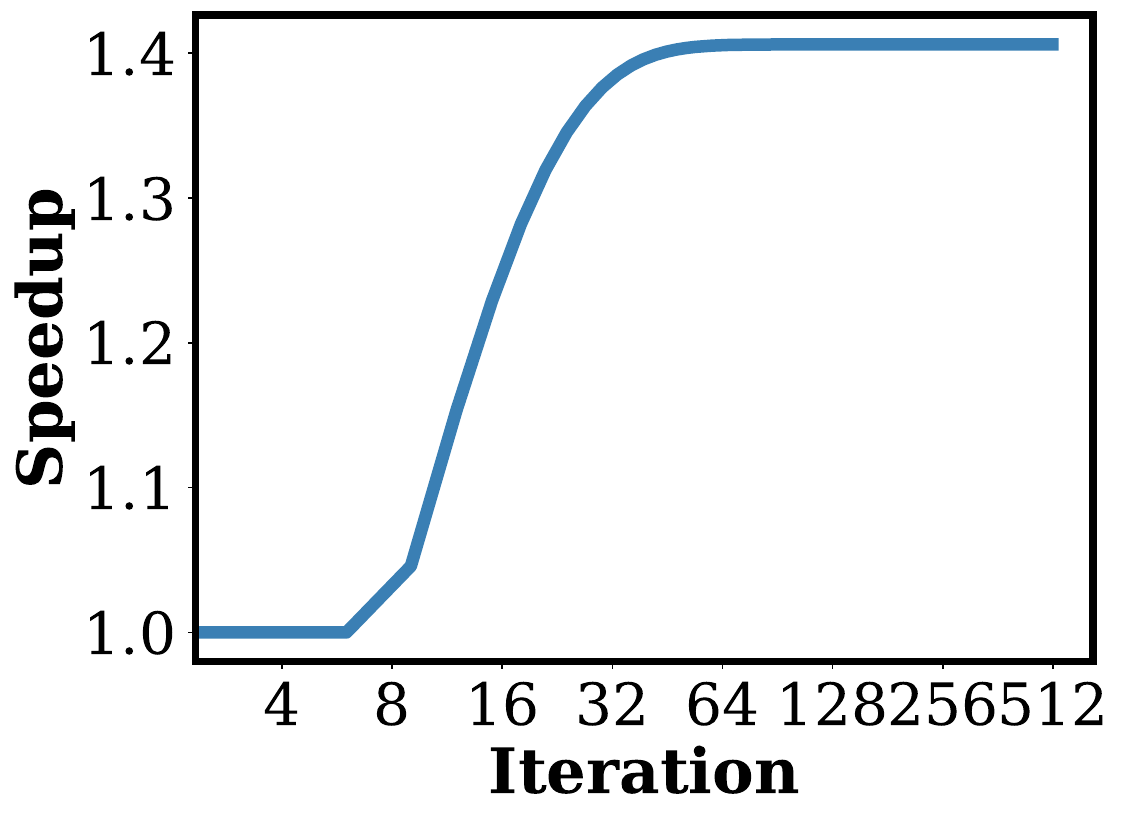}
        \captionsetup{justification=centering}
        \caption{LMDeploy DeepSeek MoE Routing}
        \label{fig:sub2}
    \end{subfigure}
    \hfill
    \begin{subfigure}[b]{0.32\textwidth}
        \centering
        \includegraphics[width=\textwidth]{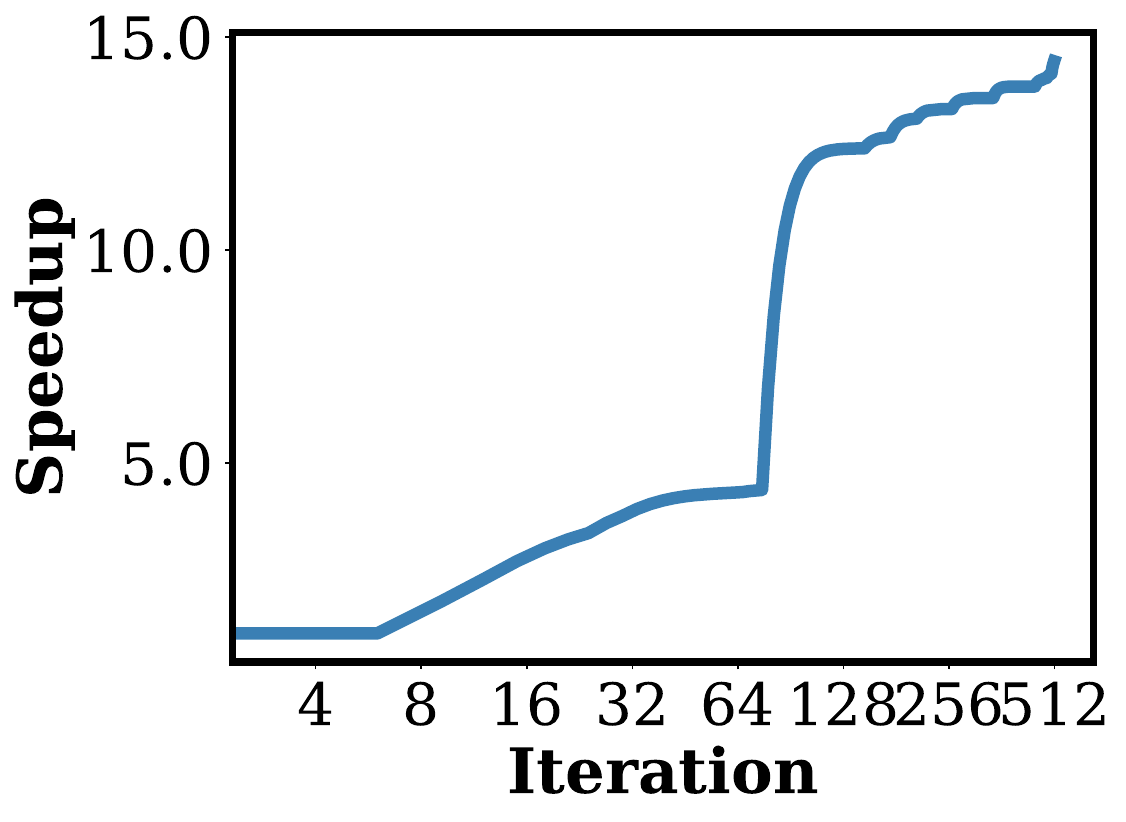}
        \captionsetup{justification=centering}
        \caption{Deepseek Engram}
        \label{fig:sub3}
    \end{subfigure}
    
    \caption{The acceleration curves during the evolutionary process of applying \modelname to operators across three real-world scenarios.}
    \label{fig:total_figure}
\end{figure}

Across these cases, the workflow is consistent. We first extract the target PyTorch module from an upstream repository and construct executable testing utilities, using LLM assistance to complement test generation when necessary. We then run \modelname to search for backend-specific kernels under compilation, correctness, and speedup feedback. Once a candidate achieves measurable performance gains, we integrate the generated kernel back into the original codebase by following the repository's pull-request guidance, and we add supplementary tests when needed. Figure \ref{fig:total_figure} illustrates the speedup growth curves associated with the deployment of our framework across three distinct scenarios.

\subsection{SGLang}

In the SGLang serving stack \cite{zheng2023sglang}, \modelname generated a fused Triton kernel for the metadata setup routine \texttt{normal\_decode\_set\_metadata}, and this kernel was subsequently merged into the FlashAttention backend.\footnote{\url{https://github.com/sgl-project/sglang/pull/20778}} The upstream patch replaces a previously unfused sequence of operations with a custom kernel that covers both a specialized fast path for the common \texttt{page\_size=1} case without sliding-window attention and a more general path supporting power-of-two page sizes with optional sliding-window attention. To make the contribution acceptable for production use, the PR also introduced dedicated unit tests across multiple batch sizes, sequence lengths, and routing settings. This case is therefore not only a microbenchmark improvement, but also a complete upstream integration exercise with implementation, validation, and deployment-facing constraints.

\begin{table}[t]
    \centering
    \caption{Isolated kernel benchmarking for the SGLang \texttt{normal\_decode\_set\_metadata} fused Triton kernel, LMDeploy fused MoE routing kernel and DeepSeek engram kernel on NV-H200.} 
    \label{tab:combined-moe}

    \begin{subtable}[t]{0.32\textwidth}
        \centering
        \captionsetup{justification=centering}
        \caption{SGLang kernel} 
        \label{tab:sglang-isolated-kernel}
        \begin{tabular}{l c}
            \toprule
            \textbf{Setting} & \textbf{Value} \\
            \midrule
            Batch size & 32 \\
            Page size & 1 \\
            Max context length & 8192 \\
            Max pool size & 1024 \\
            Seq delta & 0 \\
            Iterations & 1000 \\
            Reported speedup & $4.78\times$ \\
            \bottomrule
        \end{tabular}
    \end{subtable}
    \hfill
    \begin{subtable}[t]{0.32\textwidth}
        \centering
        \captionsetup{justification=centering}
        \caption{LMDeploy kernel} 
        \label{tab:lmdeploy-isolated-kernel}
        \begin{tabular}{l c}
            \toprule
            \textbf{Setting} & {\textbf{Value}} \\ 
            \midrule
            Batch size             & 512   \\
        Num experts            & 256   \\
        N group                & 8     \\
        Top-k group            & 4     \\
        Top-k                  & 8     \\
        Routed scaling factor  & 2.5   \\
            Reported speedup & $1.36\times$ \\
            \bottomrule
        \end{tabular}
    \end{subtable}
    \hfill
    \begin{subtable}[t]{0.32\textwidth}
        \centering
        \captionsetup{justification=centering}
        \caption{DeepSeek Engram kernel} 
        \label{tab:engram-isolated-kernel}
        \begin{tabular}{l c}
            \toprule
            \textbf{Setting} & {\textbf{Value}} \\ 
            \midrule
            Engram hidden size          & 1024   \\
            Hidden size            & 1024   \\
            Kernel size                & 4     \\
            Dilation            & 3     \\
            Hc\_mult                  & 4     \\
            Norm\_eps  & 1e-5   \\
            Reported speedup & $14.59\times$ \\
            \bottomrule
        \end{tabular}
    \end{subtable}

\end{table}


\begin{table*}[t]
    \centering
    \caption{End-to-end serving latency reported in the SGLang PR on NV-H200 using \texttt{sglang.bench\_serving} with \texttt{meta-llama/Meta-Llama-3.1-8B-Instruct}.}
    \label{tab:sglang-e2e-benchmark}
    \small
    \setlength{\tabcolsep}{7pt}
    \begin{tabular*}{\textwidth}{@{\extracolsep{\fill}} ccc S[table-format=3.2] S[table-format=2.4] S[table-format=1.2] @{}}
        \toprule
        \textbf{Max batch size} & \textbf{Input length} & \textbf{Output length} & \shortstack{\textbf{Baseline}\\\textbf{latency (ms)}} & \shortstack{\textbf{PR}\\\textbf{latency (ms)}} & \shortstack{\textbf{Rel. gain}\\\textbf{(\%)}} \\
        \midrule
        16 & 64 & 64 & 251.20 & 249.87 & 0.53 \\
        16 & 128 & 128 & 430.46 & 427.80 & 0.62 \\
        16 & 256 & 256 & 794.13 & 787.26 & 0.87 \\
        16 & 512 & 512 & 1579.02 & 1569.09 & 0.63 \\
        16 & 1024 & 1024 & 3164.77 & 3147.41 & 0.55 \\
        16 & 2048 & 2048 & 7014.71 & 6978.49 & 0.52 \\
        32 & 64 & 64 & 301.44 & 302.48 & -0.35 \\
        32 & 128 & 128 & 513.51 & 504.52 & 1.75 \\
        32 & 256 & 256 & 916.67 & 911.05 & 0.61 \\
        32 & 512 & 512 & 1761.68 & 1751.12 & 0.60 \\
        32 & 1024 & 1024 & 3674.88 & 3655.83 & 0.52 \\
        32 & 2048 & 2048 & 8274.03 & 8239.82 & 0.41 \\
        64 & 64 & 64 & 394.45 & 392.46 & 0.50 \\
        64 & 128 & 128 & 653.04 & 647.54 & 0.84 \\
        64 & 256 & 256 & 1181.10 & 1173.95 & 0.61 \\
        64 & 512 & 512 & 2337.13 & 2321.75 & 0.66 \\
        64 & 1024 & 1024 & 4827.64 & 4813.91 & 0.28 \\
        64 & 2048 & 2048 & 11689.43 & 11662.33 & 0.23 \\
        128 & 64 & 64 & 485.51 & 480.58 & 1.02 \\
        128 & 128 & 128 & 916.92 & 913.92 & 0.33 \\
        128 & 256 & 256 & 1770.41 & 1761.12 & 0.52 \\
        128 & 512 & 512 & 3353.52 & 3337.89 & 0.47 \\
        128 & 1024 & 1024 & 7480.48 & 7472.00 & 0.11 \\
        128 & 2048 & 2048 & 19039.06 & 19017.41 & 0.11 \\
        \bottomrule
    \end{tabular*}
\end{table*}

The two benchmarking tables highlight the expected gap between operator-level and system-level gains. Table~\ref{tab:sglang-isolated-kernel} reports a $4.78\times$ speedup for the fused kernel under the target configuration, showing that the metadata update routine itself benefits substantially from fusion. Table~\ref{tab:sglang-e2e-benchmark}, however, reports smaller but mostly positive latency improvements in full serving runs. This difference is expected because \texttt{normal\_decode\_set\_metadata} occupies only part of the end-to-end decoding pipeline, so even a large local optimization is diluted once scheduling, model execution, and other runtime overheads are included.

This example shows that the same search-and-integration workflow can transfer to a mature inference engine, where the bar is not just raw kernel speed but also correctness coverage, compatibility with existing execution modes, and maintainable upstream code. In that setting, sub-percent end-to-end latency reductions remain meaningful because they are obtained on a real serving path and are backed by merged production code rather than an isolated prototype.

\subsection{LMDeploy}

In LMDeploy \cite{lmdeploy2023}, we apply \modelname to the forward routing module in the MoE layer of DeepSeek-family models. The resulting implementation fuses several routing-stage operations, including sigmoid activation, bias addition, reshape, top-k selection, and masking, into a Triton kernel that was later merged into LMDeploy.\footnote{\url{https://github.com/InternLM/lmdeploy/pull/4345}} This case follows the same practical workflow as the SGLang example: \modelname searches for a backend-specific kernel, the candidate implementation is validated in the target codebase, and the final result is integrated upstream. The configurations for isolated operator benchmarking are detailed in Table \ref{tab:lmdeploy-isolated-kernel}. The merged kernel achieves more than $30\%$ speedup in isolated operator benchmarking and also improves end-to-end throughput in practical DeepSeek-v3.2 inference workloads.


\begin{table*}[t]
    \centering
    \caption{End-to-end DeepSeek-v3.2 inference throughput reported in the LMDeploy release note on NV-H200 with tensor parallelism degree 8.}
    \label{tab:lmdeploy-e2e-benchmark}
    \small
    \setlength{\tabcolsep}{4pt}
    \begin{tabular*}{\textwidth}{@{\extracolsep{\fill}}cccS[table-format=3.2]S[table-format=3.2]S[table-format=3.2]S[table-format=3.2]S[table-format=3.2]S[table-format=3.2]@{}}
        \toprule
        \multirow{2}{*}{\shortstack{\textbf{Max}\\\textbf{batch size}}} & \multirow{2}{*}{\shortstack{\textbf{Input}\\\textbf{length}}} & \multirow{2}{*}{\shortstack{\textbf{Output}\\\textbf{length}}} & \multicolumn{3}{c}{\shortstack{\textbf{Input throughput}\\\textbf{(tokens/s)}}} & \multicolumn{3}{c}{\shortstack{\textbf{Output throughput}\\\textbf{(tokens/s)}}} \\
        \cmidrule(lr){4-6} \cmidrule(lr){7-9}
        & & & \textbf{Baseline} & \shortstack{\textbf{Generated}\\\textbf{kernel}} & \shortstack{\textbf{Rel. gain}\\\textbf{(\%)}} & \textbf{Baseline} & \shortstack{\textbf{Generated}\\\textbf{kernel}} & \shortstack{\textbf{Rel. gain}\\\textbf{(\%)}} \\
        \midrule
        512 & 1K & 4K & 471.70 & 485.86 & 3.00 & 1886.78 & 1943.45 & 3.00 \\
        512 & 1K & 8K & 177.62 & 182.08 & 2.51 & 1420.97 & 1456.67 & 2.51 \\
        512 & 1K & 16K & 60.92 & 62.53 & 2.64 & 974.70 & 1000.45 & 2.64 \\
        512 & 4K & 1K & 3885.33 & 3972.52 & 2.24 & 971.33 & 993.13 & 2.24 \\
        512 & 8K & 1K & 3986.93 & 4067.89 & 2.03 & 498.37 & 508.49 & 2.03 \\
        512 & 16K & 1K & 3873.97 & 3945.79 & 1.85 & 242.12 & 246.61 & 1.85 \\
        \bottomrule
    \end{tabular*}
\end{table*}

The two tables again show the difference between local kernel improvements and end-to-end system gains. Table~\ref{tab:lmdeploy-isolated-kernel} reports about $1.36\times$ speedup for the fused routing kernel in isolated benchmarking, while Table~\ref{tab:lmdeploy-e2e-benchmark} shows consistent throughput gains of roughly $1.85\%$ to $3.00\%$ in full DeepSeek-v3.2 inference runs. This pattern is expected because the routing kernel is only one component of the full serving path, but the end-to-end improvements remain meaningful because they are measured in a realistic deployment setting.

This case further shows that \modelname can generate code with direct engineering value for a widely used inference engine rather than only for curated benchmark tasks. More broadly, it illustrates that the same search-and-integration workflow can produce upstream contributions for production systems, where practical impact depends on both kernel efficiency and successful adoption in the surrounding software stack.

\subsection{DeepSeek Engram}

Unlike the SGLang and LMDeploy cases, the Engram study does not begin from a production inference engine. Instead, we extract a recent PyTorch module from the official Engram repository,\footnote{\url{https://github.com/deepseek-ai/Engram}} which accompanies DeepSeek's conditional-memory architecture for large language models \cite{cheng2026conditional}. This setup provides a practically relevant target that is both newer than standard kernel benchmarks and closer to current research code, thereby reducing the chance that the result is driven by benchmark overlap rather than genuine generalization.


The optimized Engram implementation replaces Python-side control flow and redundant memory movement with two specialized Triton kernels that fuse gate computation, RMS normalization, depthwise convolution, and residual updates into a more streamlined execution path. By precomputing weights and storing selected intermediates in half precision, the generated version reduces both dispatch overhead and memory traffic, thereby improving GPU utilization. As shown by the Engram curve in Figure~\ref{fig:total_figure}, this case yields one of the largest gains among the three application studies: the best discovered implementation achieves a reported $14.59\times$ speedup in our local evaluation, and the resulting Engram support was later merged into DLBlas.\footnote{\url{https://github.com/DeepLink-org/DLBlas/pull/102}} The detailed configuration is given in Table~\ref{tab:engram-isolated-kernel}. This result is notable not only for its magnitude, but also because it is obtained on a fresh module drawn from a recent open-source research release rather than from a benchmark suite or a previously optimized serving stack. The Engram case therefore complements the two upstream-engine examples by showing that \modelname can transfer effectively to newly released research operators whose optimization opportunities remain largely untapped.

\section{Conclusion}

We presented Kernel-Smith, a framework for high-performance GPU kernel and operator generation that
combines a stable evaluation-driven evolutionary agent with an evolution-oriented post-training recipe. Across KernelBench, the MetaX backend, and real-world integrations such as SGLang and LMDeploy, the results show that this combination improves both search effectiveness and practical transfer beyond one-shot code generation. More broadly, these findings suggest that reliable execution feedback and
step-centric training are key ingredients for turning LLM-based kernel optimization into a usable systems
workflow. Important future directions include extending the framework to more backends, automating more of the end-to-end pull-request process for production engines, and developing more flexible agent
workflows with richer tools and adaptive search strategies.

\clearpage
\bibliographystyle{plain}
\bibliography{refs}


\clearpage
\appendix

\section{Example Evolution Prompt} \label{app:evolve-prompt-example}

To illustrate how the evolve-agent framework is instantiated for kernel generation, we provide a representative example of one evolution step below. The example is simplified for presentation, but preserves the core structure used by the system: a system prompt that specifies the optimization objective, a user prompt that injects archived programs and evaluator feedback, and a model generation that proposes the next candidate.

\definecolor{sysblue}{RGB}{25, 50, 100}      
\definecolor{usergreen}{RGB}{34, 139, 34}    
\definecolor{bglight}{RGB}{248, 249, 250}   

\newtcolorbox{sysprompt}[1]{
    breakable,
    enhanced,
    title={#1},
    colback=bglight,
    colframe=sysblue,
    coltitle=white,
    fonttitle=\bfseries\sffamily,
    attach boxed title to top left={yshift=-2mm, xshift=3mm},
    boxed title style={colback=sysblue},
    arc=2mm,           
    auto outer arc,    
    boxrule=0.8pt,
    top=8pt,
    bottom=5pt,
    shadow={1mm}{-1mm}{0mm}{black!5}
}

\newtcolorbox{userprompt}[1]{
    breakable, enhanced, title={#1},
    colback=white, colframe=usergreen, coltitle=white,
    fonttitle=\bfseries\sffamily,
    attach boxed title to top left={yshift=-2mm, xshift=3mm},
    boxed title style={colback=usergreen},
    arc=2mm, 
    auto outer arc,  
    rounded corners, boxrule=0.8pt,
    top=8pt, bottom=5pt, shadow={1mm}{-1mm}{0mm}{black!5}
}

\lstdefinestyle{promptcontent}{
    basicstyle=\ttfamily\small,
    breaklines=true,
    breakatwhitespace=false,
    columns=fullflexible,
    keepspaces=true,
    showstringspaces=false,
    extendedchars=true
}

\begin{sysprompt}{System Prompt}
\begin{lstlisting}[style=promptcontent]
You are a Triton GPU kernel optimization engineer specializing in converting PyTorch reference implementations into fast, numerically-correct Triton kernels on NVIDIA GPUs.
You are working in an evolving system that iteratively optimizes a Triton kernel.

==================== Context ====================
Input: You are given a PyTorch reference 
{reference code}
Goal: You will generate functionally equivalent but faster Triton implementations 
{custom code}

==================== Backend Device ====================
{accelerator specs}

==================== Criteria ====================
1. Compilation Pass: Make Triton code compile (performance irrelevant).
2. Numerical Correctness: Output must match PyTorch reference within float32 tolerance.
3. Performance Gain: Target more speedup than the reference code and previously generated code.

The Evolution system will pass the EVOLVE-BLOCK from the previous version.
Modify ONLY the Triton kernel source within this block - all other code is LOCKED.

==================== Constraints (MUST) ====================
NEVER change @triton.jit function signatures or parameter names.
NEVER modify grid configuration, output tensor shapes, or PID logic.
NEVER remove boundary checks or out-of-bound masks.
NEVER introduce race conditions, uninitialized reads, or sync gaps.

==================== Output Format ====================
Return ONLY the complete updated code containing the EVOLVE-BLOCK comment marker in the Markdown format. See the example below:

# ================== EVOLVE-BLOCK-START ==================
# Your code here
# =================== EVOLVE-BLOCK-END ===================

The target module class to be optimized is named with the suffix "New" after the original module name.

==================== Optimization Tips ====================
1. Minimize access to slow global memory. 
2. Kernel fusion involves combining multiple separate computational steps.
3. Maximize Occupancy. Tune BLOCK_SIZE and num_warps.
4. Grouped gemm operations: Implement Persistent Kernels.
5. Increase Arithmetic Intensity.
6. Use TMA for perfect latency hiding on NVIDIA Hopper+.
7. Setting fast_math=True allows the compiler to reorder floating-point operations.
8. Multi-Level Compilation, Tiling Hints, and Warp-Level Primitives.
9. Advanced Tiling and Work Decomposition (SplitK and SplitM).
\end{lstlisting}
\end{sysprompt}

\vspace{1em} 

\begin{userprompt}{User Prompt}
\begin{lstlisting}[style=promptcontent]
# Target Module Class
ModelNew

# Program Evolution History
## Previous Attempts
{Previous Programs}
## Top Performing Programs
{Top Performing Programs}

# Current Program
{current_program}

# Current Program Information
- Current performance metrics:
  - compiled: 1.0000
  - correctness: 0.0000
  - score: 5.0300
- Stage: runtime_error
- Runtime Error Message:
  {error message}

# Task
Rewrite the program to improve its performance on the specified metrics. Provide the complete new program code.

IMPORTANT: Make sure your rewritten program maintains the same inputs and outputs as the original program, but with improved internal implementation.
\end{lstlisting}
\end{userprompt}



\end{document}